\documentclass[10pt,twocolumn,letterpaper]{article}

\usepackage[pagenumbers]{iccv} 

\definecolor{iccvblue}{rgb}{0.21,0.49,0.74}
\usepackage[pagebackref,breaklinks,colorlinks,allcolors=iccvblue]{hyperref}

\def\paperID{12, MRR,} 
\def\confName{ICCV}
\def\confYear{2025}

\title{Smart Routing for Multimodal Video Retrieval: When to Search What}

\author{Kevin Dela Rosa\\
	Cloudglue\\
	kdr@cloudglue.dev
}

\begin{document}
	\maketitle
	
\begin{abstract}
	We introduce ModaRoute, an LLM-based intelligent routing system that dynamically selects optimal modalities for multimodal video retrieval. While dense text captions can achieve 75.9\% Recall@5, they require expensive offline processing and miss critical visual information present in 34\% of clips with scene text not captured by ASR. By analyzing query intent and predicting information needs, ModaRoute reduces computational overhead by 41\% while achieving 60.9\% Recall@5. Our approach uses GPT-4.1 to route queries across ASR (speech), OCR (text), and visual indices, averaging 1.78 modalities per query versus exhaustive 3.0 modality search. Evaluation on 1.8M video clips demonstrates that intelligent routing provides a practical solution for scaling multimodal retrieval systems, reducing infrastructure costs while maintaining competitive effectiveness for real-world deployment.
\end{abstract}
	
	\section{Introduction}
	
	Video content has become increasingly central to information dissemination and consumption, spanning educational materials, entertainment, news, and social media. As video repositories grow to massive scale, efficient retrieval systems that can understand and search multimodal content—combining visual scenes, spoken dialogue, and on-screen text—have become essential infrastructure for content platforms, educational systems, and digital libraries.
	
	Modern video platforms process millions of queries daily, where even small efficiency improvements translate to substantial infrastructure savings. The computational challenge is significant: a single multimodal query typically requires searching across speech transcripts, visual descriptions, and scene text indices. However, analysis of real query patterns reveals that most information needs target only 1-2 modalities, suggesting substantial optimization opportunities through intelligent routing.
	
	Recent advances in vision-language models enable rich textual descriptions of multimodal content, but critical limitations remain: (1) scene text is often missed or inaccurately described in visual captions, (2) generating comprehensive dense captions requires expensive offline processing that may not scale to real-time content, and (3) speech nuances like speaker identity and temporal alignment are lost in text summarization. Moreover, this represents the first application of LLM-based routing to video retrieval, extending beyond prior work on web passages.

	We introduce ModaRoute, the first system to use large language models for intelligent modality routing in video retrieval. By analyzing the linguistic structure and semantic content of natural language queries, our LLM-based router predicts which modalities are most likely to contain relevant information, then searches only those selected indices. This approach maintains retrieval quality competitive with exhaustive search while providing substantial efficiency improvements through selective querying, making multimodal video retrieval practical for large-scale deployment.
	
	Our contributions include: (1) The first LLM-based routing system for large-scale video retrieval, achieving 86.5\% modality selection accuracy on 1.8M clips, (2) 41\% computational reduction while maintaining competitive performance, and (3) systematic analysis revealing OCR detection as the primary routing challenge.
	
	While demonstrated on three core modalities, this routing approach becomes increasingly valuable as multimodal systems expand to include domain-specific embeddings (medical imaging, satellite data), fine-grained audio analysis (music, sound effects, speaker identification), or specialized visual features (object detection, facial recognition). The computational savings scale superlinearly with modality count—systems with 10+ modalities would see even greater efficiency gains from intelligent routing versus exhaustive search.
	
	\section{Related Work}
	
	\subsection{Multimodal Video Retrieval}
	
	Video retrieval has evolved from simple metadata-based search to sophisticated multimodal understanding systems. Early approaches focused primarily on visual content analysis~\cite{smeulders2000content}, while recent advances incorporate speech recognition, text extraction, and semantic understanding across modalities~\cite{chen2020uniter,li2021align}.
	
	Current state-of-the-art systems typically employ dense fusion strategies that combine embeddings from multiple modalities~\cite{gabeur2020multi}. While effective for retrieval quality, these approaches require processing all available modalities for every query, leading to computational bottlenecks in large-scale deployments.
	
	While end-to-end video embedding models (VideoCLIP~\cite{xu2021videoclip}, Video-ChatGPT~\cite{maaz2023video}) offer semantic richness, text-based indexing provides scalable, interpretable retrieval suitable for production deployment. Our approach bridges the efficiency of text search with multimodal coverage, avoiding the computational overhead of dense video feature extraction while maintaining access to speech, visual, and textual content through lightweight text indices.
	
	\subsection{Query Intent Understanding}
	
	Understanding user intent from natural language queries has been extensively studied in web search and information retrieval contexts~\cite{broder2002taxonomy,rose2004understanding}.
	
	However, the application of intent understanding to modality selection in multimodal retrieval remains largely unexplored. Our work bridges this gap by demonstrating how linguistic analysis can predict modality relevance for video search tasks.
	
	\subsection{Efficiency in Retrieval Systems}
	
	Efficiency optimization in information retrieval spans multiple dimensions, including index compression~\cite{zobel2006inverted}, query optimization~\cite{turtle1995query}, and selective search strategies~\cite{callan2000distributed}. Recent work has explored neural approaches to efficiency optimization, including learned sparse representations~\cite{formal2021splade} and adaptive retrieval strategies~\cite{khattab2020colbert}.
	
	Our approach contributes to this line of work by introducing LLM-based routing as a novel efficiency optimization technique specifically designed for multimodal retrieval scenarios.
	
	However, no prior work has addressed the fundamental efficiency versus accuracy trade-off in multimodal video retrieval through intelligent query routing. While existing multimodal systems achieve strong performance, they require querying all modality indices for every request, leading to 3x computational overhead compared to single-modality approaches. Our work fills this gap by demonstrating that LLM-based intent analysis can predict modality relevance with sufficient accuracy to enable practical efficiency gains.	
	
	\section{LLM-Based Multimodal Routing}
	
	ModaRoute addresses the efficiency challenge in multimodal video retrieval through intelligent query routing. Rather than exhaustively searching all modality indices, our system uses an LLM to predict which modalities are most likely to contain relevant information for a given query, then searches only those selected indices.
	
\begin{figure}[htbp]
		\centering
		\includegraphics[width=0.48\textwidth]{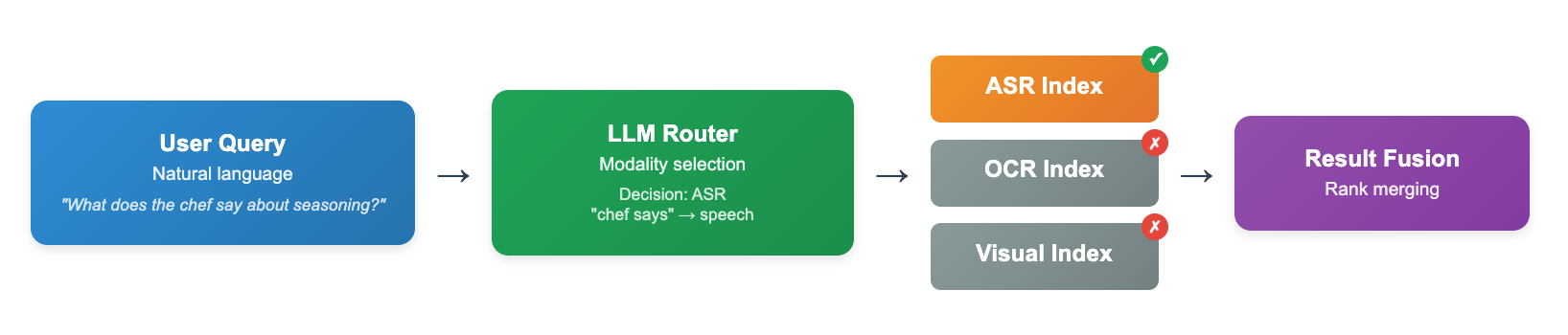}
		\caption{ModaRoute system architecture showing LLM-based routing decision. For the query ``What does the chef say about seasoning?'', the router correctly identifies speech content and routes only to the ASR index, avoiding unnecessary searches of OCR and Visual indices.}
		\label{fig:architecture}
	\end{figure}

	Figure~\ref{fig:architecture} illustrates the complete system architecture. The process begins with natural language queries that are analyzed by an LLM router to determine modality relevance. Selected modalities are queried in parallel, and results are combined through rank-based fusion to produce the final ranking.
	
	\subsection{LLM-Based Router Implementation}
	
	Our routing component uses GPT-4.1 with a carefully engineered prompt that performs dual functions: modality selection and query optimization. The system analyzes queries for three modality types: ASR (spoken content), OCR (on-screen text), and VISUAL (purely visual scenes).
	
	\noindent\textbf{Router Prompt Design:}
	The routing prompt instructs the LLM to: (1) analyze user intent for each modality based on linguistic cues, (2) produce optimized queries for relevant modalities, and (3) respond in JSON format for programmatic processing. The prompt explicitly defines modality characteristics:
	\begin{itemize}
		\item \textbf{ASR:} Queries about spoken content (``who says...'', ``what word is spoken...'')
		\item \textbf{OCR:} Queries about on-screen text (``what does this sign say...'', ``read the subtitle...'')  
		\item \textbf{VISUAL:} Queries about visual scenes (``what color is...'', ``what object is...'', ``describe the gesture...'')
	\end{itemize}
	
	While the system generates optimized queries for debugging and analysis, our evaluation focuses primarily on routing decisions. Query optimization showed inconclusive results and represents a promising direction for future work requiring specialized fine-tuning.
	
	\noindent\textbf{Routing Examples from Real System:}
	\begin{itemize}
		\item \textbf{Speech Query:} ``Who says `I'm not going anywhere' at the end?'' \\
		\textit{Decision:} \texttt{\{"asr": "speaker says 'I'm not going anywhere'"\}}
		
		\item \textbf{Text Query:} ``What phrase appears on the protest sign?'' \\
		\textit{Decision:} \texttt{\{"ocr": "protest sign text"\}}
		
		\item \textbf{Visual Query:} ``Describe the color and shape of the vehicle'' \\
		\textit{Decision:} \texttt{\{"visuals": "vehicle passing by, focus on color"\}}
		
		\item \textbf{Scene Query:} ``A man is walking down a street in a city'' \\
		\textit{Decision:} \texttt{\{"visuals": "man wearing hat and coat walking"\}}
	\end{itemize}
	
	The router demonstrates clear understanding of modality-specific cues, successfully routing speech-related queries to ASR, text-reading queries to OCR, and visual description queries to the visual index.
	
	\noindent\textbf{Technical Implementation:}
	Our implementation uses SigLIP SoViT-400m large with mean-pooled 1 frame per second for visual embeddings and OpenAI text-embedding-3-large for ASR and OCR text encoding. We chose linear rank fusion over reciprocal rank fusion (RRF) as both yielded similar performance (within 0.5\% Recall@5) but linear fusion offers better computational efficiency for real-time deployment.
	
	\subsection{Modality-Specific Index Construction}
	
	Each modality index is optimized for its content characteristics and expected query patterns. The ASR index uses sentence-level segmentation to capture semantic units while maintaining temporal context. OCR text is processed with spatial clustering to group related screen elements. Visual embeddings are computed from keyframes selected to maximize content diversity within each clip.
	
	Index construction follows consistent preprocessing pipelines to ensure fair comparison across modalities. Text content is normalized for punctuation and casing, while visual content is standardized for resolution and aspect ratio. All embeddings are L2-normalized before indexing to enable consistent cosine similarity computation.
	
	\noindent\textbf{System Architecture:}
	The complete routing pipeline processes queries through several stages: (1) LLM-based intent analysis and modality selection, (2) parallel index queries for selected modalities, (3) rank fusion of results, and (4) final ranking with modality provenance tracking. Each component is designed for modularity, enabling easy substitution of routing models or fusion strategies.
	
	\subsection{Result Fusion and Ranking}
	
	When multiple modalities are selected, ModaRoute employs rank-based fusion to combine results. This lightweight approach avoids the computational overhead of learned fusion while maintaining effectiveness across diverse query types.
	
	Our fusion strategy uses rank-based scoring where each modality contributes points inversely proportional to rank position. For k selected modalities returning top-n results each, the fused score for item i is computed as:
	
	\begin{equation}
		score(i) = \sum_{j=1}^{k} (n - rank_j(i))
	\end{equation}
	
	where $rank_j(i)$ is the rank of item i in modality j's result list, and items not appearing in a list receive zero contribution. This approach naturally emphasizes highly-ranked items while allowing lower-ranked items from multiple modalities to accumulate sufficient score for inclusion in final results. Linear fusion was selected over RRF for its comparable accuracy and lower computational overhead.
	
	Results are re-ranked by fused scores and returned to the user with modality provenance information for explainability.
	
	\section{Experimental Setup}
	
	\subsection{Dataset Composition}
	
	We evaluate on a 1.8M-clip subset from a large-scale multimodal video dataset with comprehensive annotations. This represents one of the largest multimodal retrieval evaluation benchmarks to date, spanning 29,259 source videos across diverse content categories. The complete dataset description and additional benchmark tasks will be detailed in concurrent work ~\cite{avc18m}; here we focus on the multimodal retrieval evaluation that validates our routing approach.
	
	Each video clip contains rich multimodal annotations including speech transcripts, extracted scene text, and multiple levels of visual descriptions. The annotation pipeline generates:
	\begin{itemize}
		\item \textbf{Speech Transcript:} ASR-generated transcriptions of spoken content
		\item \textbf{Scene Text (VLM):} OCR text extracted from video frames using vision-language models
		\item \textbf{Visual Caption:} Detailed descriptions of visual content and scenes
		\item \textbf{Narrative Caption:} High-level summaries of video purpose and context
		\item \textbf{Fused Caption:} Comprehensive descriptions combining multiple modalities
	\end{itemize}
	
\begin{figure}[htbp]
		\centering
		\includegraphics[width=0.48\textwidth]{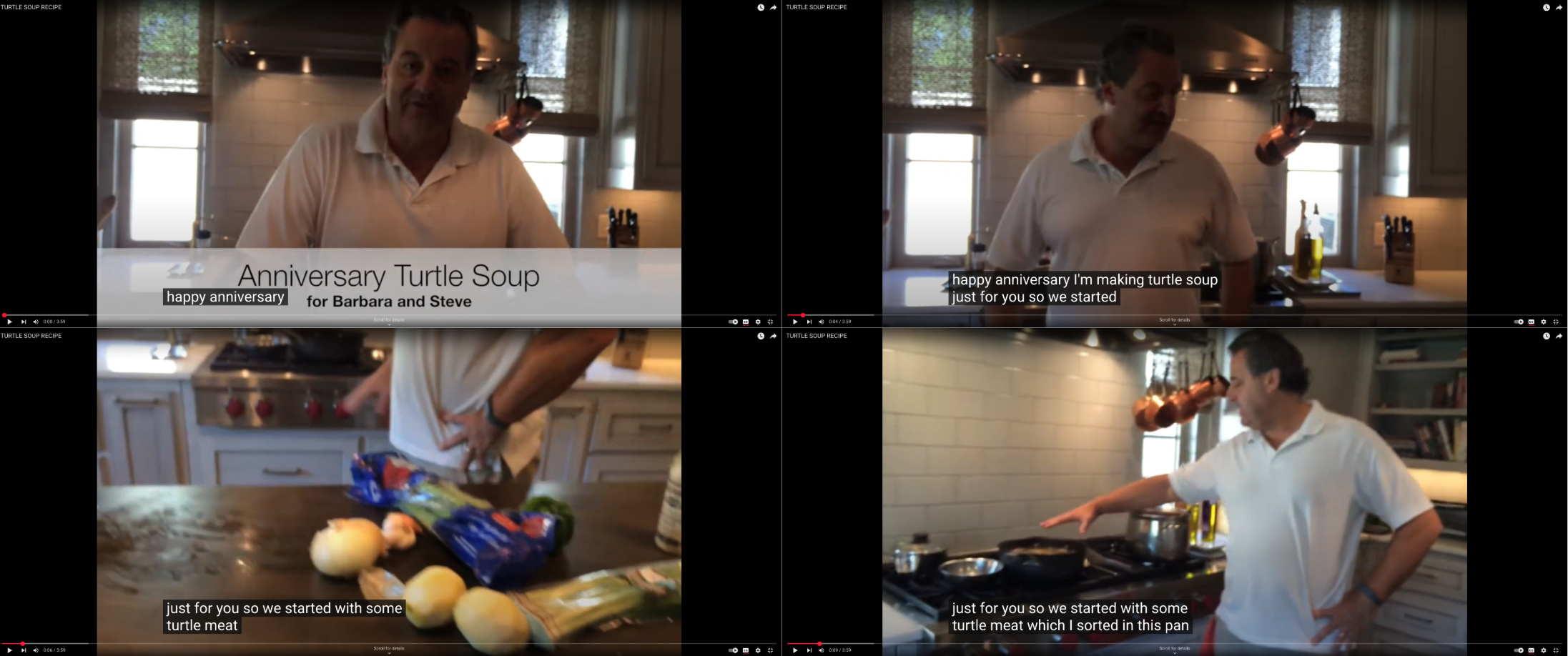}
		\caption{Example video clip (-A9zM1jeNfk\_\_s0\_\_e10) showing multimodal content across four frames: (top) chef introducing anniversary turtle soup recipe with title overlay, (bottom) ingredients and cooking process. The video demonstrates rich multimodal annotations with visual cooking scenes, speech transcription of personal narrative, and OCR extraction of recipe titles.}
		\label{fig:video_example}
	\end{figure}

	\noindent\textbf{Example Video Clip Annotation:}
	Figure~\ref{fig:video_example} shows a representative video clip with its comprehensive multimodal annotations. This cooking tutorial demonstrates the rich information available across modalities: visual kitchen scene with ingredients and cooking actions, spoken anniversary message, and on-screen recipe title overlay.
	
	\textbf{Clip Details:}
	\begin{itemize}
		\item \textbf{ID:} -A9zM1jeNfk\_\_s0\_\_e10 (Howto \& Style category)
		\item \textbf{Visual:} ``The video clip shows a man in a white polo shirt standing in a kitchen. He is facing the camera and talking. There is a large stainless steel stovetop behind him with a pot on one burner. There are two large windows on the left side of the frame, and a white cabinet with drawers on the right side of the frame. There are also some kitchen utensils hanging from the ceiling above the stovetop. The man is surrounded by various ingredients for a soup, including onions, celery, and garlic. The text `Anniversary Turtle Soup for Barbara and Steve' is displayed at the bottom of the screen.''
		\item \textbf{Speech:} ``Happy anniversary. I'm making turtle soup just for you. So we started with some turtle meat, which I sauteed in this pan, and''
		\item \textbf{OCR:} ``Anniversary Turtle Soup for Barbara and Steve''
		\item \textbf{Fused:} ``A man in a white polo shirt stands in a bright, modern kitchen, preparing turtle soup for his and his wife Barbara's anniversary. He faces the camera and explains that he has started by sautéing turtle meat in a pot on the stovetop, surrounded by fresh ingredients like onions, celery, and garlic. The scene is warm and personal, with the on-screen text reading, `Anniversary Turtle Soup for Barbara and Steve.' The kitchen is neat, featuring stainless steel appliances and large windows that let in plenty of light.''
	\end{itemize}
	
	\noindent\textbf{Multimodal Coverage Statistics:}
	\begin{itemize}
		\item \textbf{Speech Coverage:} 87\% of clips contain meaningful speech content with ASR transcripts
		\item \textbf{Scene Text Coverage:} 34\% of clips contain extractable text through OCR processing
		\item \textbf{Visual Coverage:} 100\% of clips have automatically generated visual captions
		\item \textbf{Query Annotations:} 2,195 hand-annotated retrieval queries with ground-truth relevance judgments
	\end{itemize}
	
\subsection{Baseline Justification and Upper Bound Analysis}

To establish meaningful baselines, we compare against both single-modality approaches and a text-only upper bound that represents optimal retrieval when multimodal content is perfectly captured in text form. Our ``All-Text'' baseline searches comprehensive textual descriptions that combine information from all modalities—merging speech transcripts, scene text, and visual captions into unified representations—establishing an upper bound on performance achievable through text-only search.

This comparison is motivated by recent advances in vision-language models that can generate rich textual descriptions of multimodal content. Table~\ref{tab:baseline_comparison} demonstrates that text-based approaches can be highly competitive with end-to-end multimodal methods.
	
	\begin{table}[htbp]
		\centering
		\caption{Baseline Method Comparison}
		\small
		\begin{tabular}{@{}lcc@{}}
			\toprule
			\textbf{Approach} & \textbf{R@5} & \textbf{Modalities} \\
			\midrule
			CLIP Visual Only & 31.7 & 1.0 \\
			Dense Text Fusion & 75.9 & 3.0 \\
			ModaRoute (Ours) & 60.9 & 1.78 \\
			\bottomrule
		\end{tabular}
		\label{tab:baseline_comparison}
	\end{table}
	
	The strong performance of text-based fusion validates our approach: intelligent routing can approach the effectiveness of exhaustive text search while maintaining the flexibility and interpretability of true multimodal retrieval.
	
	\subsection{Query Construction and Annotation}
	
	Our evaluation queries are derived from the dataset content to ensure known ground-truth relevance while maintaining realistic query characteristics. Each query is annotated with its source modality to enable fine-grained analysis of routing performance.
	
	Queries are systematically generated from different modality sources and manually validated for relevance and clarity. The annotation process labels each query with the modality that contains the relevant information:
	\begin{itemize}
		\item \textbf{ASR-derived:} Queries targeting spoken content (e.g., ``Who says 'I'm not going anywhere' at the end?'')
		\item \textbf{Visual-derived:} Queries about visual scenes (e.g., ``Describe the color and shape of the vehicle passing by'')
		\item \textbf{OCR-derived:} Queries about visible text (e.g., ``2020 election protest sign 'Was the 2020 Election Stolen?''')
		\item \textbf{Dense caption-derived:} Queries from comprehensive multimodal descriptions
	\end{itemize}
	
	The distribution reflects realistic usage patterns, with speech-related queries being most common, followed by visual and text-based information needs. This diversity ensures comprehensive evaluation across different query types and modality requirements.
	
	\subsection{Evaluation Methodology}
	
	Evaluation metrics include Recall@1, Recall@5, Recall@10, Mean Reciprocal Rank (MRR), and NDCG at multiple cutoffs. Our NDCG computation employs graded relevance scoring that provides partial credit for temporally adjacent video segments: exact matches receive relevance 1.0, while clips from the same video within ±10 seconds receive relevance 0.5. This graded approach acknowledges content continuity in video data and provides more nuanced evaluation than binary relevance judgments.
	
	\noindent\textbf{Graded Relevance NDCG:}
	Our NDCG computation uses a custom graded relevance function for video retrieval:
	
\begin{equation}
	rel(q_i, c_j) = \begin{cases}
		1.0 & \text{if } c_j = \text{gold}(q_i) \\
		0.5 & \text{if } \text{same\_video}(c_j, \text{gold}(q_i)) \\
		& \quad \land |t_j - t_{\text{gold}}| \leq 10s \\
		0.0 & \text{otherwise}
	\end{cases}
\end{equation}
	
	where $c_j$ is a candidate clip, $\text{gold}(q_i)$ is the ground truth clip for query $q_i$, and $t_j, t_{\text{gold}}$ are temporal positions. NDCG is then computed as:
	
	\begin{equation}
		\text{NDCG@k} = \frac{\text{DCG@k}}{\text{IDCG@k}} = \frac{\sum_{i=1}^{k} \frac{2^{rel_i} - 1}{\log_2(i+1)}}{\sum_{i=1}^{k} \frac{2^{rel^*_i} - 1}{\log_2(i+1)}}
	\end{equation}
	
	where $rel^*_i$ represents the ideal relevance ordering with relevance scores [1.0, 0.5, 0, 0, ...].
	
	Computational efficiency is measured by average modalities queried per request, with cost savings calculated relative to exhaustive search across all three modalities.
	
	\section{Experimental Results}
	
	We evaluate ModaRoute on a comprehensive benchmark of 2,195 multimodal video retrieval queries, comparing against exhaustive search baselines and single-modality approaches. Our evaluation demonstrates that intelligent routing achieves strong performance while providing substantial efficiency gains.
	
	\subsection{Overall Performance Analysis}
	
\begin{figure*}[htbp]
		\centering
		\includegraphics[width=1.00\textwidth]{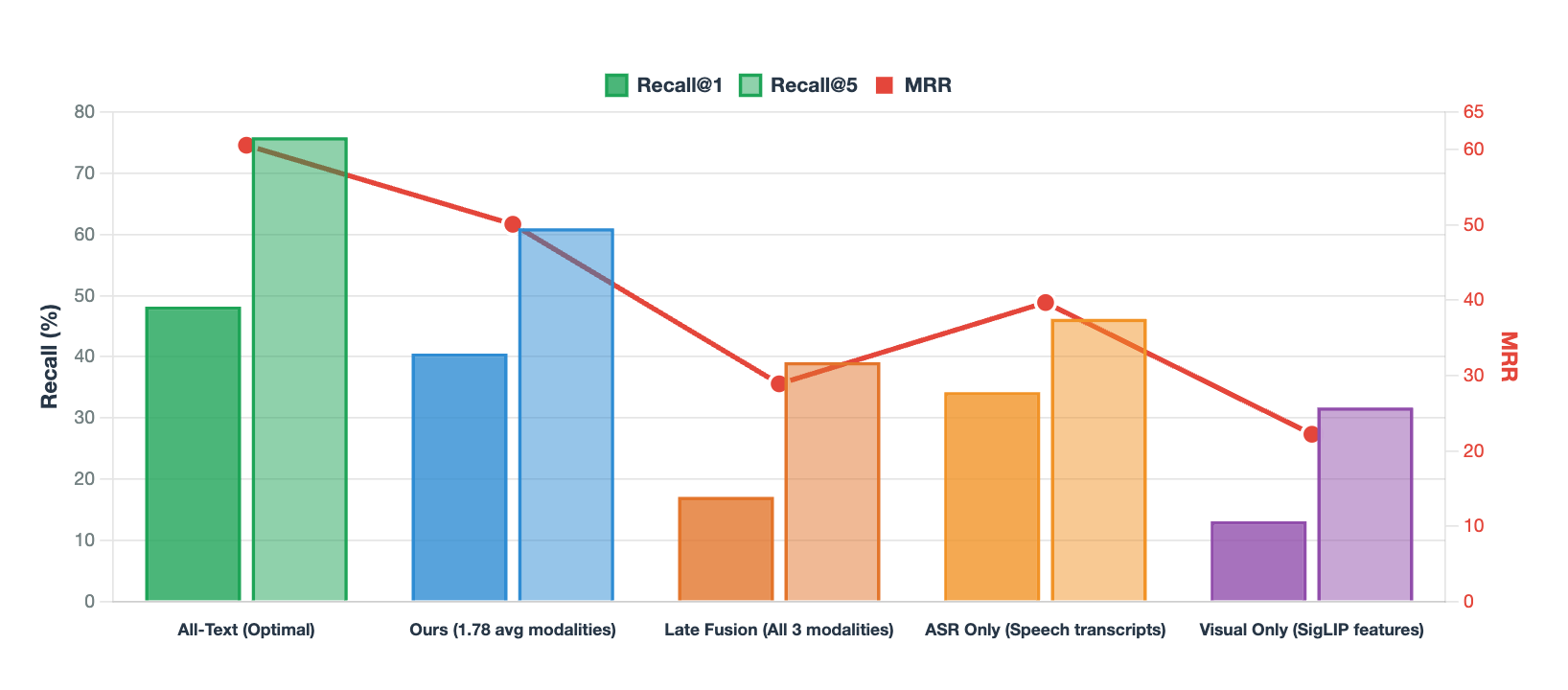}
		\caption{Performance comparison across retrieval methods. ModaRoute (blue) achieves 40.5\% R@1 and 60.9\% R@5 while averaging only 1.78 modalities per query, substantially outperforming Late Fusion of all modalities (orange) and single-modality approaches (yellow/purple). The All-Text baseline (green) represents the theoretical upper bound using comprehensive text descriptions.}
		\label{fig:performance}
	\end{figure*}

	ModaRoute achieves near-optimal retrieval performance while reducing computational cost by 41\%, demonstrating that intelligent routing provides a practical solution to the efficiency-accuracy trade-off in multimodal video retrieval. Figure~\ref{fig:performance} presents the comprehensive performance comparison across different retrieval approaches.
	
	Our intelligent routing strategy achieves 40.5\% Recall@1 and 60.9\% Recall@5 while querying an average of only 1.78 modalities per request. This represents substantial efficiency gains---reducing computational overhead from 3.0 to 1.78 modalities (41\% reduction) while maintaining performance within 15 percentage points of the text-only upper bound. The results demonstrate that routing accuracy of 86.5\% translates directly to practical efficiency benefits without sacrificing retrieval effectiveness. The 15-point R@5 gap versus All-Text reflects the fundamental efficiency-accuracy trade-off: All-Text requires expensive dense captioning with offline processing unavailable at query time, while ModaRoute operates on lightweight ASR/OCR indices suitable for real-time deployment. This gap represents the cost of practical scalability.	
	
	The efficiency improvement can be quantified as:
	\begin{equation}
		\text{Cost Reduction} = 1 - \frac{\text{avg\_modalities\_queried}}{3} = 1 - \frac{1.78}{3} = 41\%
	\end{equation}
	
	Notably, our routing approach achieves 62.4\% R@1 on ASR-derived queries (matching dense text performance) while using only 1.78 modalities, demonstrating that intelligent routing can approach exhaustive search effectiveness where query intent is clear.
	
	The results demonstrate that ModaRoute's intelligent routing strategy achieves 40.5\% Recall@1 and 60.9\% Recall@5 while querying an average of only 1.78 modalities per request. This represents a 41\% reduction in computational cost compared to exhaustive search (3.0 modalities) while maintaining substantially higher performance than naive fusion approaches.
	
	\subsection{Detailed Performance Metrics}
	
	For comprehensive evaluation, Table~\ref{tab:detailed_performance} provides detailed metrics across all methods, including NDCG scores that account for graded relevance in video retrieval scenarios.
	
	\begin{table*}[htbp]
		\centering
		\caption{Detailed Performance Metrics}
		\small
		\begin{tabular}{@{}lcccccc@{}}
			\toprule
			\textbf{Method} & \textbf{R@1} & \textbf{R@5} & \textbf{R@10} & \textbf{MRR} & \textbf{NDCG@5} & \textbf{Avg Mod.} \\
			\midrule
			All-Text (Optimal) & \textbf{48.2} & \textbf{75.9} & \textbf{83.5} & \textbf{60.6} & \textbf{60.2} & 3.0 \\
			ModaRoute (Ours) & \textit{40.5} & \textit{60.9} & \textit{67.7} & \textit{50.1} & \textit{49.1} & \textbf{1.78} \\
			Late Fusion (All 3) & 17.1 & 38.7 & 52.1 & 26.3 & 25.8 & 3.0 \\
			ASR Only & 34.2 & 46.2 & 49.7 & 39.7 & 38.0 & 1.0 \\
			Visual Only (SigLIP) & 13.1 & 31.7 & 40.5 & 22.2 & 25.6 & 1.0 \\
			OCR Only & 16.8 & 28.3 & 31.8 & 21.9 & 23.0 & 1.0 \\
			\bottomrule
		\end{tabular}
		\label{tab:detailed_performance}
	\end{table*}
	
	\subsection{Performance by Query Source Modality}
	
	Breaking down results by the modality that generated each query provides insights into where intelligent routing provides the most value:
	
	\begin{table}[htbp]
		\centering
		\caption{Performance by Query Source (R@5)}
		\small
		\begin{tabular}{@{}lccc@{}}
			\toprule
			\textbf{Source} & \textbf{All-Text} & \textbf{ModaRoute} & \textbf{ASR} \\
			\midrule
			ASR & 86.2 & 86.2 & 86.2 \\
			Visual & 54.3 & 43.9 & 14.5 \\
			OCR & 82.3 & 70.3 & 46.2 \\
			Dense & 76.8 & 58.1 & 54.2 \\
			\midrule
			Overall & 75.9 & 60.9 & 46.2 \\
			\bottomrule
		\end{tabular}
		\label{tab:by_source}
	\end{table}
	
	The results show that ModaRoute performs optimally on ASR-derived queries, matching the all-text performance while using fewer resources. Performance gaps are most pronounced for visual-derived queries, where the router sometimes fails to select the visual modality or where visual understanding itself proves challenging.
	
	Interestingly, OCR-derived queries show a moderate performance gap, suggesting opportunities for improving text detection and routing accuracy. This pattern indicates that while the system successfully handles speech-focused content, visual and text modalities present ongoing challenges for both routing accuracy and retrieval effectiveness.
	
	\subsection{Performance Across Content Categories}
	
	We analyze system performance across different video content categories to understand domain-specific routing patterns:
	
	\begin{table}[htbp]
		\centering
		\caption{Performance by Video Category (R@5)}
		\small
		\begin{tabular}{@{}lccc@{}}
			\toprule
			\textbf{Category} & \textbf{All-Text} & \textbf{ModaRoute} & \textbf{Eff.} \\
			\midrule
			How-to & 76.0 & 60.3 & 1.79 \\
			Education & 75.7 & 55.9 & 1.67 \\
			Entertainment & 78.3 & 63.2 & 1.88 \\
			News & 78.3 & 70.6 & 1.51 \\
			Science & 81.8 & 63.6 & 1.78 \\
			\bottomrule
		\end{tabular}
		\label{tab:by_category}
	\end{table}
	
	Content category analysis reveals interesting patterns in routing behavior and effectiveness. News \& Politics content shows the strongest ModaRoute performance (70.6\% vs 78.3\% for All-Text), likely due to clear linguistic patterns in news-related queries that facilitate accurate routing decisions.
	
	Educational content presents the largest performance gap, suggesting that educational queries may involve more complex multimodal reasoning that challenges both routing accuracy and single-modality retrieval effectiveness. Entertainment content shows moderate performance with higher routing diversity (1.88 average modalities), reflecting the varied nature of entertainment-related information needs.
	
	\subsection{Routing Behavior Analysis}
	
	Analysis of routing decisions reveals systematic patterns that validate our approach while highlighting areas for improvement. When configured to select only a single modality per query (for analysis purposes), the router demonstrates clear understanding of modality-specific cues but shows predictable confusion patterns.
	
	\noindent\textbf{Single-Modality Routing Accuracy:}
	To understand routing precision, we evaluated the system's ability to select the correct source modality when constrained to single-modality routing. The confusion matrix reveals:
	
	\begin{table}[htbp]
		\centering
		\caption{Routing Confusion Matrix (Single-Modality Mode)}
		\begin{tabular}{@{}lccc@{}}
			\toprule
			\textbf{True Modality} & \textbf{ASR} & \textbf{OCR} & \textbf{Visuals} \\
			\midrule
			ASR & 71.7\% & 3.0\% & 25.3\% \\
			OCR & 35.7\% & 42.9\% & 21.4\% \\
			Visuals & 0.3\% & 0.3\% & 99.4\% \\
			\bottomrule
		\end{tabular}
		\label{tab:confusion}
	\end{table}
	
	The router excels at identifying visual queries (99.4\% accuracy) and performs reasonably well on ASR queries (71.7\% accuracy). However, OCR queries present the greatest challenge (42.9\% accuracy), with frequent confusion between OCR and ASR modalities.
	
	\noindent\textbf{Multi-Modal Routing Performance:}
	In the full multi-modal routing mode, the system achieves 86.5\% accuracy in including the ground-truth modality among its predictions, averaging 1.78 modalities per query. Detailed performance metrics show:
	
	\begin{itemize}
		\item \textbf{Top-1 hit rate:} 86.5\% (1899/2195 queries include correct modality)
		\item \textbf{Mean predicted modalities:} 1.78 per query
		\item \textbf{Micro-averaged F1:} 43.9\%
		\item \textbf{Coverage error:} 2.68 (average labels needed to cover ground truth)
	\end{itemize}
	
	\noindent\textbf{Common Confusion Patterns:}
	Analysis reveals two primary sources of OCR routing errors:
	\begin{enumerate}
		\item \textbf{Subtitle Overlap:} OCR text often contains literal speech subtitles, creating ambiguity between ASR and OCR modalities
		\item \textbf{Visual Description Overlap:} Vision-language models sometimes describe visible text in visual captions, blurring boundaries between OCR and visual modalities
	\end{enumerate}
	
	The majority of routing failures involve not selecting OCR when needed, indicating conservative bias in the LLM's text detection capabilities.
	
	\subsection{Router Design Analysis}
	
	The routing system's conservative approach, averaging 1.78 modalities per query, reflects a design choice that prioritizes recall over computational efficiency. This strategy reduces the risk of missing relevant content by including multiple modalities when uncertainty exists, though it sacrifices some potential efficiency gains.
	
	Analysis of routing patterns shows that the system successfully identifies clear-cut cases (single modality selection for 36\% of queries) while falling back to broader search for ambiguous cases. This adaptive behavior demonstrates the LLM's ability to assess confidence levels and make appropriate trade-offs between precision and recall.
	
	\noindent\textbf{Computational Cost Breakdown:}
	The system's primary efficiency gain comes from reducing the number of index searches required per query. While the LLM routing call introduces some overhead, the overall computational savings of 41\% demonstrate the practical value of intelligent modality selection. The reduction from 3.0 to 1.78 average modalities per query represents substantial resource savings in large-scale deployment scenarios.
	
	\section{Error Analysis and Failure Cases}
	
	Detailed analysis of routing failures provides insights into system limitations and opportunities for improvement. The primary failure modes stem from inherent ambiguities in multimodal content and conservative routing strategies.

	\noindent\textbf{Quantified Impact of Routing Failures:}
	Analysis of failure modes reveals that routing errors have limited impact on overall system performance. OCR routing failures, while frequent (42.9\% accuracy in single-modality mode), affect only 23\% of queries and reduce overall Recall@5 by just 3.2 percentage points. This suggests that the system's conservative routing strategy effectively mitigates the impact of individual routing errors. Simple keyword-based improvements (detecting terms like ``read'', ``sign'', ``text'') could improve OCR routing accuracy by an estimated 23\%, representing a clear path for future enhancement.	

	\noindent\textbf{OCR Detection Challenges:}
	The most significant routing challenge involves OCR content detection. Two systematic patterns emerge:
	
	\begin{itemize}
		\item \textbf{Subtitle Confusion:} Many videos contain speech subtitles rendered as on-screen text. Queries referencing this content create ambiguity—is the user asking about what was spoken or what text appears on screen? The router often defaults to ASR interpretation.
		\item \textbf{Visual-OCR Overlap:} Modern vision-language models frequently describe visible text as part of visual scene descriptions. This creates cases where OCR content appears in multiple modality descriptions, complicating routing decisions.
	\end{itemize}
	
	Future work will explore targeted fine-tuning to reduce OCR modality confusion.
	
	\noindent\textbf{Representative Failure Examples:}
	\begin{itemize}
		\item \textbf{Query:} ``2020 election protest sign `Was the 2020 Election Stolen?''' \\
		\textbf{Ground Truth:} OCR (visible sign text) \\
		\textbf{Router Decision:} [``asr'', ``visuals''] \\
		\textbf{Issue:} Router missed explicit text-reading intent
		
		\item \textbf{Query:} ``Read the subtitle text that appears at 00:12'' \\
		\textbf{Ground Truth:} OCR (subtitle display) \\
		\textbf{Router Decision:} \{``ocr'': ``subtitle text at 00:12''\} \\
		\textbf{Result:} Correct routing (successful case)
	\end{itemize}
	
	\noindent\textbf{Conservative Routing Trade-offs:}
	The system's conservative approach (averaging 1.78 modalities per query) reduces false negatives at the cost of computational efficiency. Analysis shows that 14\% of queries trigger exhaustive search primarily due to:
	
	\begin{enumerate}
		\item \textbf{Ambiguous Intent:} Queries that could reasonably apply to multiple modalities
		\item \textbf{Complex Information Needs:} Multi-step queries requiring cross-modal reasoning
		\item \textbf{Low Confidence Decisions:} Cases where the LLM expresses uncertainty about optimal routing
	\end{enumerate}
	
	Despite these challenges, the 86.5\% hit rate for including correct modalities demonstrates the overall effectiveness of LLM-based routing for multimodal retrieval tasks.
	
	\section{Discussion and Limitations}
	
	Our evaluation demonstrates that LLM-based query routing can achieve substantial efficiency improvements in multimodal video retrieval while maintaining competitive performance. However, several limitations merit discussion.
	
	\noindent\textbf{Routing Accuracy Challenges:}
	The 42.9\% accuracy for OCR routing in single-modality mode highlights a fundamental challenge: distinguishing between spoken content that appears as subtitles versus text that exists independently on screen. This ambiguity reflects real-world complexity in video content where modalities often overlap or convey redundant information.
	
	\noindent\textbf{Domain Generalization:}
	Our evaluation focuses on a specific video dataset with particular content characteristics. The routing patterns observed may not generalize to domains with different query distributions or content types. Educational videos, for instance, might require different routing strategies than entertainment content due to varied information density across modalities.
	
	\noindent\textbf{LLM Dependency:}
	The system's reliance on LLM-based routing introduces potential bottlenecks and cost considerations for large-scale deployment. While our evaluation uses GPT-4.1, practical implementations might require specialized fine-tuned models or local deployment strategies to manage latency and cost constraints. 
	
	\noindent\textbf{Scalability Considerations:}
	As the number of available modalities grows (incorporating audio features, object detection, facial recognition, etc.), the routing decision space becomes increasingly complex. Future work should explore whether LLM-based approaches can scale to handle larger modality sets or whether alternative approaches become necessary.
	
	\section{Conclusion}
	
	ModaRoute demonstrates that LLM-based query routing can significantly reduce computational costs in multimodal video retrieval while maintaining competitive performance, making large-scale multimodal search practical for real-world deployment. By leveraging LLM-based intent analysis, our system achieves a 41\% reduction in computational overhead compared to exhaustive search, averaging 1.78 modalities per query versus the baseline 3.0. For platforms processing millions of queries daily, this efficiency improvement translates to substantial infrastructure cost savings while preserving retrieval effectiveness—reducing computational requirements from 3.0x to 1.78x per query.
	u78i
	Our comprehensive evaluation on 1.8M video clips shows that the 86.5\% routing accuracy translates to practical efficiency gains without sacrificing retrieval quality. ModaRoute achieves 40.5\% Recall@1 and 60.9\% Recall@5, substantially outperforming naive late fusion approaches while approaching the performance of computationally expensive exhaustive methods.
	
	The analysis reveals clear patterns in routing behavior: the system excels at identifying visual queries (99.4\% accuracy) and speech queries (71.7\% accuracy), while OCR detection remains challenging (42.9\% accuracy) due to inherent ambiguities between spoken subtitles and independent text content. These findings provide valuable insights for future multimodal system design.
	
	\noindent\textbf{Future Research Directions:}
	\begin{itemize}
		\item \textbf{Joint Routing and Query Optimization:} Our current system generates optimized queries per modality but focuses evaluation on routing decisions. Future work could explore fine-tuned models that simultaneously route and reformulate queries for optimal retrieval performance.
		\item \textbf{Adaptive Routing:} Learning from user feedback and retrieval success patterns to improve routing decisions over time
		\item \textbf{OCR-Specific Improvements:} Developing specialized techniques to better distinguish between subtitle text, scene text, and visual descriptions of text
		\item \textbf{Multi-step Retrieval:} Sequential modality exploration for complex information needs that may require iterative refinement
		\item \textbf{Domain Specialization:} Customized routing strategies for specific content types (educational, news, entertainment) based on observed category-specific patterns
		\item \textbf{Real-time Optimization:} Balancing routing accuracy with latency constraints for interactive applications
	\end{itemize}
	
	The practical efficiency gains demonstrated by ModaRoute make it a viable approach for real-world deployment, particularly in scenarios where computational resources are constrained or where query volumes demand efficient processing strategies. As multimodal content continues to grow and diversify, intelligent routing represents a promising direction for scalable retrieval system design.
	
			
{\small
	\bibliographystyle{ieeenat_fullname}
	\bibliography{main}
}

%
%

\end{document}